\documentclass{article}
\usepackage{spconf,amsmath,graphicx}

\usepackage{times}
\usepackage{epsfig}
\usepackage{graphicx}
\usepackage{amsmath}
\usepackage{amssymb}

\usepackage{times}
\usepackage{epsfig}
\usepackage{graphicx}
\usepackage{amsmath}
\usepackage{amssymb}

\usepackage[noend]{algpseudocode}
\usepackage{booktabs}
\usepackage{multirow}
\usepackage{algorithmicx,algorithm}
\usepackage{pifont}
\usepackage{array}
\usepackage{enumitem}
\usepackage{soul}
\usepackage{color}
\usepackage[colorlinks,linkcolor=red,anchorcolor=blue,citecolor=green]{hyperref}

\newcommand{\PreserveBackslash}[1]{\let\temp=\\#1\let\\=\temp}
\newcolumntype{C}[1]{>{\PreserveBackslash\centering}p{#1}}
\newcolumntype{R}[1]{>{\PreserveBackslash\raggedleft}p{#1}}
\newcolumntype{L}[1]{>{\PreserveBackslash\raggedright}p{#1}}

\title{Enhancing Prototypical Few-Shot Learning by Leveraging the Local-Level Strategy}
\name{Junying~Huang,~Fan~Chen,~Keze~Wang,~Liang~Lin,~Dongyu~Zhang*\thanks{*Corresponding author}}
\address{Sun Yat-Sen University}
\begin{document}
%
\maketitle
\begin{abstract}
Aiming at recognizing the samples from novel categories with few reference samples, few-shot learning (FSL) is a challenging problem. We found that the existing works often build their few-shot model based on the image-level feature by mixing all local-level features, which leads to the discriminative location bias and information loss in local details. To tackle the problem, this paper returns the perspective to the local-level feature and proposes a series of local-level strategies. Specifically, we present (a) a local-agnostic training strategy to avoid the discriminative location bias between the base and novel categories, (b) a novel local-level similarity measure to capture the accurate comparison between local-level features, and (c) a local-level knowledge transfer that can synthesize different knowledge transfers from the base category according to different location features. Extensive experiments justify that our proposed local-level strategies can significantly boost the performance and achieve 2.8$\%$–7.2$\%$ improvements over the baseline across different benchmark datasets, which also achieves the state-of-the-art accuracy.
\end{abstract}
\section{Introduction}
\label{sec:intro}
Few-shot learning aims to recognize samples from novel categories with only a few labeled samples. Existing methods can be roughly divided into two main streams: (a) Metric-based approaches \cite{hou2019cross, kye2020transductive, snell2017prototypical, vinyals2016matching, zhang2020deepemd,wang2019simpleshot, multi-pro} focus on learning an appropriate feature space for both base and novel categories, and then classify the query data according to their distance between support data on the feature space. ProtoNet \cite{snell2017prototypical} proposes to build a prototype for each category and recognize by the nearest neighbor rule. The recent FEAT \cite{ye2020set} also proposes to refine the prototype by a Transformer module, and DeepEMD proposes a high-complexity distance. (b) Optimization-based approaches \cite{finn2017model, hu2020empirical, lee2019meta, gidaris2018dynamic, liu2019large} focus on improving the transferability of the fully trained model and transferring the knowledge of base categories to novel categories. For example, \cite{gidaris2018dynamic, liu2019large} propose an attention module to refine query features/prototypes by the knowledge of the base category.

\begin{figure}[t]
	\begin{center}
		\includegraphics[width=\columnwidth]{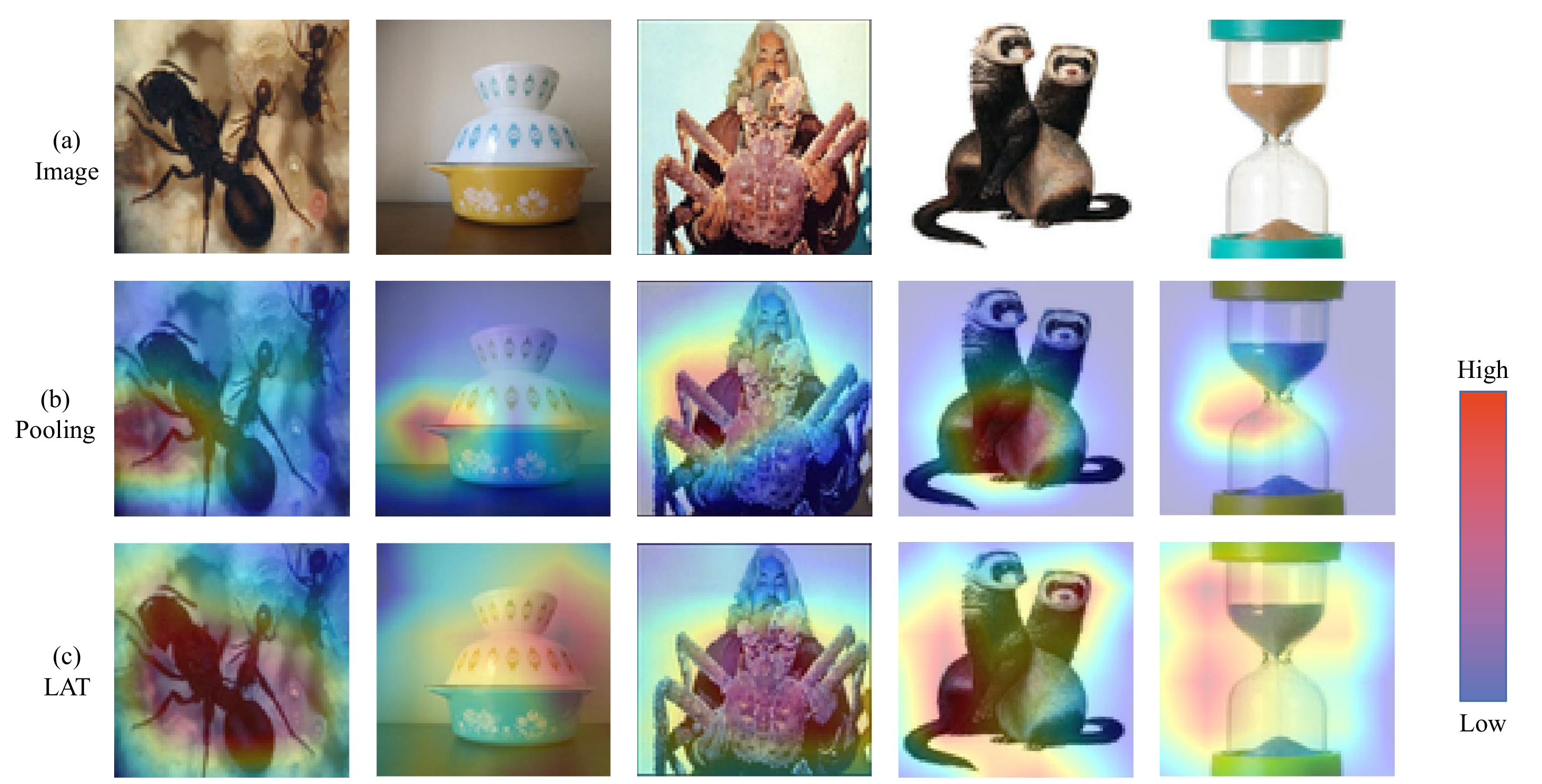}
	\end{center}
	\vspace{-15pt}
	\caption{Feature heat map of novel category samples extracted from models with global pooling or local-agnostic training (LAT): The color overlaid on the original image represents the $L_2$-norm of the location feature (Red means high and blue means low). The higher $L_2$-norm can represent more model attention to the corresponding location because the corresponding feature has greater weight for the classification.}
	\label{fig: heatmap}
	\vspace{-10pt}
\end{figure}

Although these methods have achieved encouraging success, we observe that they often follow the common routine in the traditional image classification task, i.g., extracting the image-level feature vector by global pooling \cite{lin2013network} and then building their models on the image-level feature space. Under the fully supervised training by a large amount of labeled data, global pooling can improve the performance by driving the model to focus attention on the local features considered discriminative and suppress the features on the insignificant location. For example, in Fig. \ref{fig: heatmap} (b), the high-weight local features are only distributed at concentrated locations. However, these discriminative features are distributed in different locations for different categories. We call this difference \textit{discriminative location bias}. Few-shot learning requires the model to recognize the object belonging to the novel category. Hence, the discriminative location bias will cause the model to over-pay attention to the location beneficial to the base category while ignoring the discriminative features of the novel category. As shown in Fig. \ref{fig: heatmap} (b), the model trained with global pooling only pays attention to the insignificant location on the novel category images, resulting in the wrong classification.

To avoid the discriminative location bias, we first propose a local-level classification loss and a local-level regularization loss to realize local-agnostic training (LAT), which forces the model to pay attention to all locations instead of only the discriminative location. As shown in Fig. \ref{fig: heatmap} (c), the high-weight local features from the model trained with the LAT strategy can cover the complete object of novel categories.

Furthermore, during the inference phase, the image-level features of the novel category are either misled by the discriminative location bias (from the model trained with global pooling) or confused by too much information (from the model trained with LAT strategy). Meanwhile, the image-level feature vector also destroys the local details in the image. Therefore, we further propose two local-level strategies during inference for a comprehensive understanding of novel category images. First, we propose a hybrid local-level similarity measure (LSM) to measure the similarity between local-level features accurately, composed of local distance and matching distance. Then, we propose a local-level knowledge transfer (LKT), which can synthesize the different knowledge transfers from base categories for different location features.

Extensive experiments justify that our proposed local-level strategies can significantly boost the few-shot performance. On the ProtoNet \cite{snell2017prototypical} baseline, it achieve 2.8$\%$–7.2$\%$ improvements. Compared with the competitive works, it doesn't require the complex algorithm \cite{zhang2020deepemd} or additional model \cite{multi-pro, ye2020set} but still achieves the state-of-the-art accuracy.

\section{Background Knowledge}
\label{sec:preliminary}

Given disjoint category sets $\{\mathcal C_{base}$, $\mathcal C_{val}$, $\mathcal C_{novel}\}$ and the corresponding datasets $\{\mathcal D_{base}$, $\mathcal D_{val}$, $\mathcal D_{novel}\}$, the objective of few-shot learning (FSL) is to train a robust model on $\mathcal D_{base}$ which can generalize well on $\mathcal D_{novel}$ with few reference samples per category. The reference data is called the support set and the recognized data is called the query set. Generally, the model is evaluated by the $N$-way $K$-shot episode classification \cite{vinyals2016matching} in FSL. In each episode, a support set $\mathcal S$ and a query set $\mathcal Q$ are sampled from $\mathcal D_{novel}$ with the same $N$ categories, where $\mathcal S$ contains $K$ examples for each category with available labels. Then, the evaluation metric is defined as the average accuracy on $\mathcal Q$ under multiple episodes. In this work, we use the setting in \cite{vinyals2016matching}, i.e., $N = 5$, $K \in \{1, 5\}$ and $\mathcal Q$ randomly samples 15 examples for each category.

We build our model on top of ProtoNet \cite{snell2017prototypical}, which is a popular metric-based method due to its simplicity and effectiveness.
In each episode, the ProtoNet first computes a prototype (center) for each class in $\mathcal S$,
\begin{equation}
    \label{eq:pro_int}
    P_c = \frac{1}{|\mathcal{S}_c|} \sum_{I \in \mathcal S_c} f_{\Theta}(I)
\end{equation}
where $f_{\Theta}$ means the embedding extractor, $\mathcal{S}_c$ means the subset of $\mathcal S$ with label $c$ and $|\cdot|$ means the set size.
After that, the probability distribution of the query image $I_q$ is predicted by the distance function $D$ and softmax function $\sigma$,
\begin{equation}
    \label{eq:pred}
    q(I_q) = \sigma([-D(I_q, P_c)]_{c=1}^N)
\end{equation}

\begin{figure}[h]
    \begin{center}
    \includegraphics[width=\columnwidth]{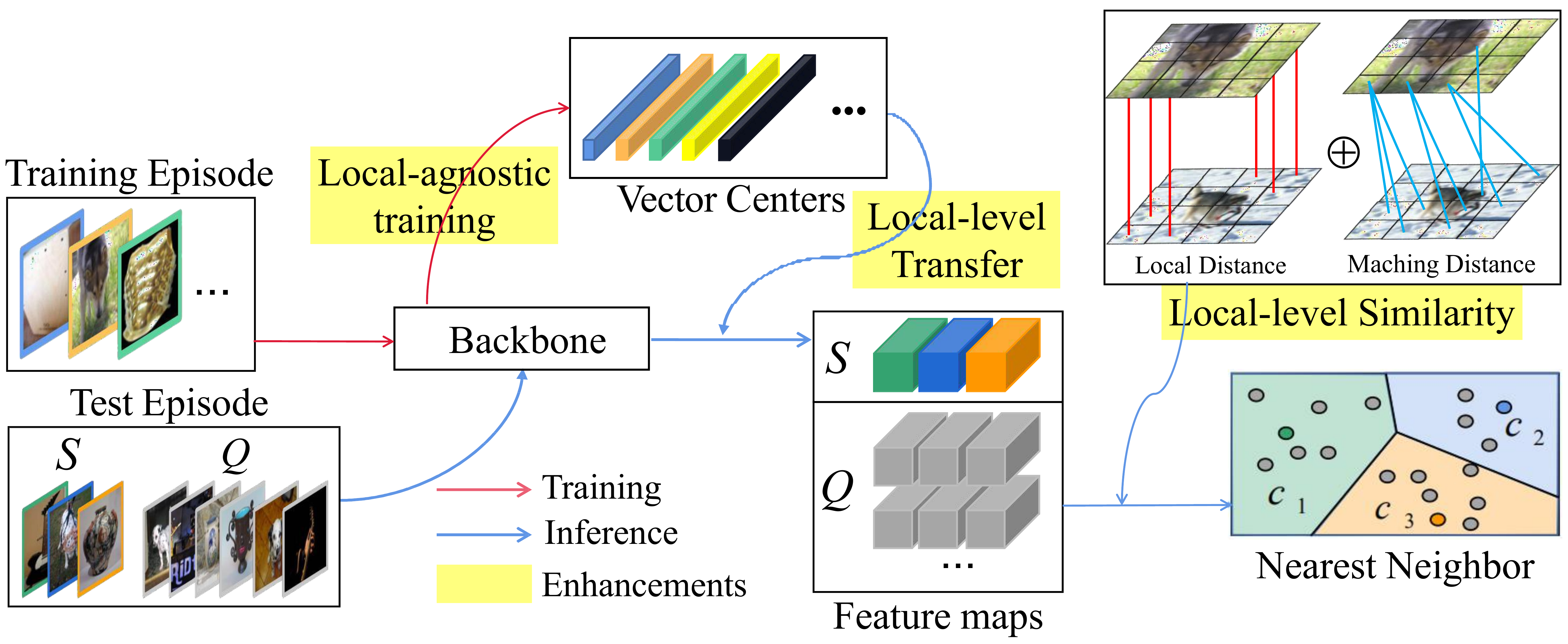}
    \end{center}
    \vspace{-15pt}
  \caption{Overview of framework: The red and blue arrows mean the process in the training and inference phase respectively. We also mark the proposed enhancement in yellow.
    }
    \vspace{-10pt}
\label{fig:overview}
\end{figure}

\section{Methodology}
\label{approach}
In this section, we describe the model architecture with local-agnostic training objective (Sec. \ref{sec:inference}), the local-level similarity measure (Sec. \ref{sec:distance}), and the local-level knowledge transfer (Sec. \ref{sec:adaption}) in order.
The overview of our framework is depicted in Fig. \ref{fig:overview}. In particular, we mark the position of three proposed strategies in the framework by yellow, and the local-agnostic training and local-level knowledge transfer are visualized in Fig. \ref{fig:loss} and Fig. \ref{fig:adaption} in detail.

\subsection{Framework}
\label{sec:inference}
\noindent\textbf{Model Architecture}: Our model consists of an embedding extractor $\phi$ and a classifier $f_W$. Specifically, we use ResNet-12 \cite{he2016deep} with four blocks as the extractor and extract the feature map from the last block $\phi^{(4)}(I)$ for classification. The classifier $f_W: R^{d\times w\times h} \rightarrow R^{c\times w\times h}$ consists of a 1x1 convolution function $W\in R^{d\times c}$ without bias and a softmax function $\sigma$, which classifies each local feature on the last feature map for the local-level classification:
\begin{equation}
    f_W(\phi^{(4)}(I)) = \left [\sigma([W_l^T \phi^{(4)}_{i, j}(I)]_{l=1}^{c})\right ]_{i=1, j=1}^{w, h}
\end{equation}

\begin{figure}[t]
	\begin{center}
	    \vspace{-10pt}
		\includegraphics[width=\columnwidth]{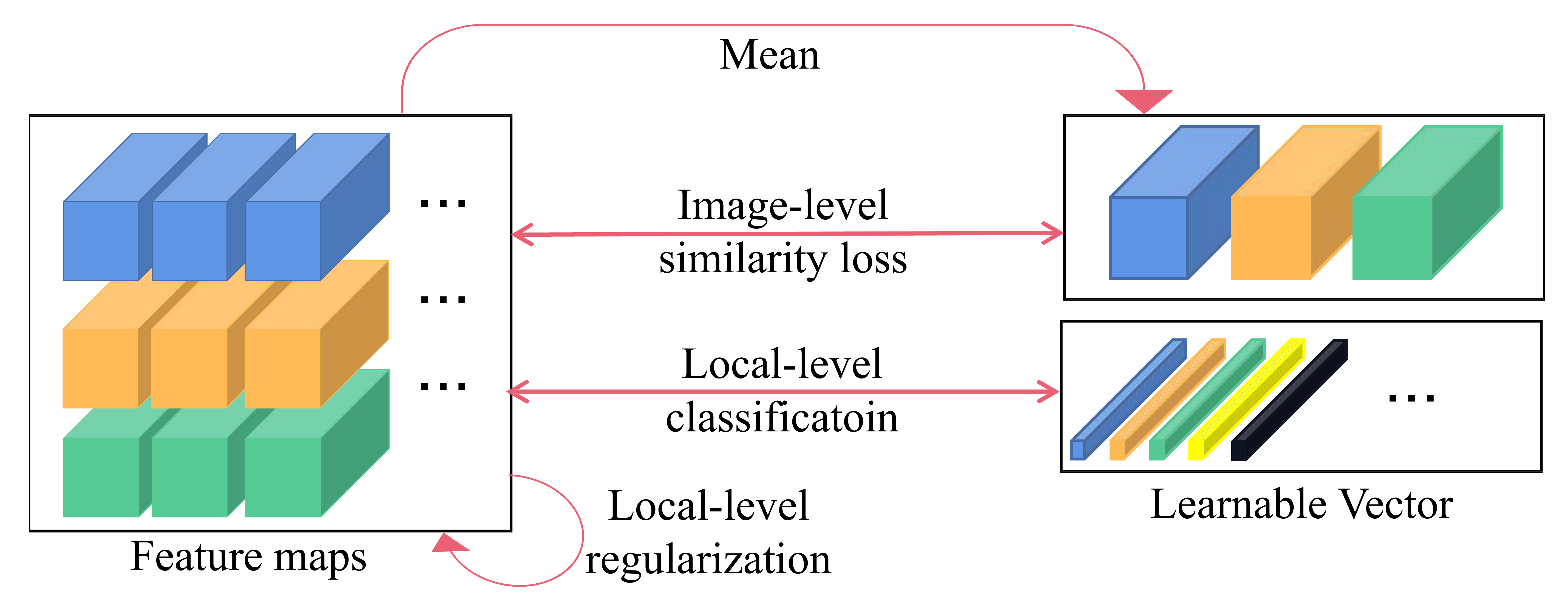}
	\end{center}
	\vspace{-15pt}
	\caption{ The complete training objective, in which the learnable vectors are the weights of the 1x1 convolution classifier. }
	\label{fig:loss}
	\vspace{-8pt}
\end{figure}

\noindent\textbf{Training Objective}: 
We adopt the episode-based training \cite{snell2017prototypical} to avoid the over-fitting on $\mathcal D_{base}$. In each training, an $N'$-way $K'$-shot episode $E$ that is considered as both support set and query set was sampled. As shown in Fig. \ref{fig:loss}, the training objective consists of the image-level similarity loss $L_S$, as well as the local-level classification $L_C$ and regularization loss $L_R$ for local-agnostic training. The image-level similarity loss aims to minimize the intra-class distance and maximize the inter-class distance in the feature space, which is defined as
\begin{equation}
    \label{eq:l_g}
    \mathcal L_S(E, \phi) = \sum_{(I, y) \in E} l(\sigma( [-D(\phi^{(4)}(I),\ \ P_c) ]_{c=1}^{N'} ),\ \ y)
\end{equation}
where $l$ is cross-entropy function, $\sigma$ is softmax function, the $P_c$ is the prototype of category $c$ computed by Eq. \ref{eq:pro_int}, and the distance function $D$ is described in Sec. \ref{sec:distance}.

Then we consider local-agnostic training: In the image-level classification loss, the gradient of low-discrimination locations is diluted by the global pooling function, which leads to the discriminative location bias. To avoid the problem, we first replace the image-level classification loss with the local-level classification loss (Eq. \ref{eq: lc}). It optimizes the classification loss for each location on the feature map, which encourages the model to focus on all location features, whether high or low discrimination. Specifically, we use a 1x1 convolution function described above as the classifier, and the local-level classification loss is computed by the cross-entropy $l$:
\begin{equation}
    \label{eq: lc}
    \mathcal L_C(E, \phi, W) = \sum_{(I, y)\in E} \sum_{\substack{i=1\\ j=1}}^{w, h} l(f_W^{(i, j)}(\phi^{(4)}(I)), y)
\end{equation}

Then, we propose a local-level regularization loss $L_R$, which further avoids the discriminative location bias by punishing the model for paying too much attention to some location features. Specifically, we adopt the $L_2$-norm of each local feature to evaluate the attention of the model to this location. Thus the local-level regularization loss is defined as

\begin{equation}
    \mathcal L_R(E, \phi) = \sum_{(I, y)\in E} Var\ \{ || \phi_{(i, j)}^{(4)}(I) ||_2 \}_{i=1, j=1}^{w, h}
\end{equation}
where $Var$ means the variance across all local features. The final optimization objective is the sum of three losses:
\begin{equation}
    \label{eq:l_all}
    J = \mathcal L_C(E, \phi, W) + \lambda_S \mathcal L_S(E, \phi) + \lambda_R \mathcal L_R(E, \phi)
\end{equation}

\subsection{Local-level Similarity Measure}
\label{sec:distance}
Different from the image-level feature, the local-level feature is structural, so the similarity between local-level features should also consider their spatial structures. DeepEMD \cite{zhang2020deepemd} proposed to compute the minimal spatial matching between local-level features, but this destroys the original spatial structure. MetaOptNet \cite{lee2019meta} and MCT \cite{kye2020transductive} directly compare the spatial similarity, but it may not be accurate when the object posture and position vary. Different from them, we consider both the original spatial structures and the minimal spatial matching, which are visualized in Fig. \ref{fig:overview}. For each feature map $x \in R^{d\times w\times h}$, after normalizing them by Eq. (\ref{eq:normalize}), we first directly compare the spatial similarity of local-level features between query sample and prototype $p$ by local distance $d_L$:
\begin{equation}
    \label{eq:normalize}
    \mathcal N(x) = \frac{x}{||x||_F}
\end{equation}
\begin{equation}
    \label{eq:d_g}
    d_{L}(x, p) = \left\|  (\mathcal N(x) - \mathcal N(p) ) \right\|_F^2
\end{equation}
Then we propose a simple minimum matching distance $d_M$ between local-level features to capture the correct comparison when object position and posture vary, which is defined as
\begin{equation}
    \label{eq:d_m}
    d_{M}(x, p) = \sum_{\substack{i=1\\ j=1}}^{w, h} \min_{a, b} \left\| ( \mathcal N_{i, j}(x) - \mathcal N_{a, b}(p) ) \right\|^2_2 ,
\end{equation}
The final distance between the feature map and prototype is defined as the weighted sum of the two distances: 
\begin{equation}
    D(x, p) = d_{L}(x, p) + \gamma\ d_{M}(x, p)
\end{equation}

\begin{figure}[h]
	\begin{center}
        \vspace{-10pt}
	    \includegraphics[width=0.7\columnwidth]{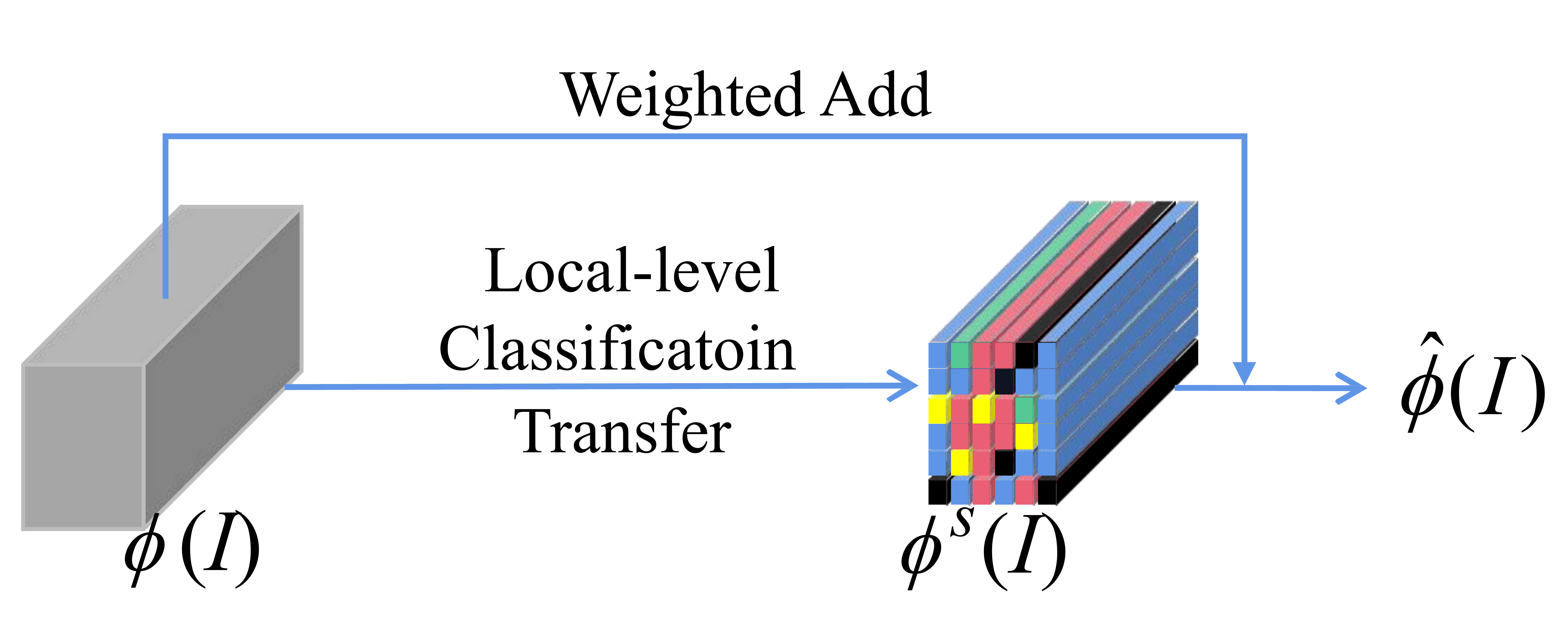}
	\end{center}
	\vspace{-15pt}
	\caption{ The process of LKT. ``Transfer'' means to replace each local feature as the weighted sum of the base knowledge. }
	\label{fig:adaption}
	\vspace{-10pt}
\end{figure}

\subsection{Local-level Knowledge Transfer}
\label{sec:adaption}
In FSL, the semantic information of novel datasets is unavailable during training, which leads the distribution of the novel category samples difficult to converge in the feature space. \cite{gidaris2018dynamic, liu2019large} propose an attention module to refine the distribution of novel samples by the similarity with the base category. Different from them, we argue that the image-level feature cannot correctly describe the semantic information of the novel category sample (Sec. \ref{sec:intro}).
Thus we introduce the local-level knowledge transfer during inference that refines all the location features separately. Specifically, we adopt the convergent classifier weight $W$ (described in Sec. \ref{sec:inference}) as the knowledge (prototype) of base categories. As shown in Fig. \ref{fig:adaption}, for each sample from $\mathcal D_{novel}$, we first compute a $\mathcal D_{base}$ similar-version $\phi^{s}(I)$ for the feature map $\phi^{(4)}(I)$, on which each location feature is computed by the probability-weighted sum of $W$, 
\begin{equation}
    \phi^{s}_{i, j}(I) =   \sum_{l=1}^{|W|}  [f_W^{(i, j)}(\phi^{(4)}(I))]_l \cdot W_l 
\end{equation}
Then, the local-level feature is refined as the weighted sum of the similar-version and original-version after normalization:
\begin{equation}
    \label{eq:adaption}
    \hat{\phi}_{i,j}^{(4)}(I) = \beta \mathcal N_{i, j}(\phi^{(4)}(I)) + (1-\beta) \mathcal N_{i, j} (\phi^{s}(I))
\end{equation}


\begin{table*}[htp]
    \small
    \setlength{\abovecaptionskip}{0.0cm}
    \setlength{\abovecaptionskip}{0.0cm}
    \setlength{\belowcaptionskip}{0.1cm}
    \setlength{\tabcolsep}{0.8mm}{
      \centering
        \begin{tabular}{cccC{2.0cm}C{2.0cm}C{2.0cm}C{2.0cm}C{2.0cm}C{2.0cm}}
        \toprule
        \multirow{2}{*}{Model} & \multirow{2}{*}{Backbone} & \multicolumn{2}{c}{MiniImageNet}  & \multicolumn{2}{c}{TieredImageNet} & \multicolumn{2}{c}{Cifar-FS}  \\
        \cline{3-8}
        & & 1-shot & 5-shot & 1-shot & 5-shot & 1-shot & 5-shot  \\
        \midrule
        MetaOptNet-SVM \cite{lee2019meta} & ResNet-12 
        & 62.64 $\pm$ 0.61 & 78.63 $\pm$ 0.46 & 65.99 $\pm$ 0.72 & 81.56 $\pm$ 0.53 & 72.00 $\pm$ 0.70 & 84.20 $\pm$ 0.50        \\
        Simple-Shot \cite{wang2019simpleshot} & WRN-28-10
        & 63.50 $\pm$ 0.20 & 80.33 $\pm$ 0.14 & 69.75 $\pm$ 0.20 & 85.50 $\pm$ 0.14 &       ---        &       ---        \\
        CAN \cite{hou2019cross} & ResNet-12
        & 63.85 $\pm$ 0.48 & 79.44 $\pm$ 0.34 & 69.89 $\pm$ 0.51 & 84.23 $\pm$ 0.37 &       ---        &       ---       \\
        R2-D2+Aug \cite{liu2020task} & ResNet-12
        & 64.79 $\pm$ 0.45 & 81.08 $\pm$ 0.32 &       ---        &       ---       & 76.51 $\pm$ 0.47 & 87.63 $\pm$ 0.34    \\
        S2M2$\_$R \cite{mangla2020charting} & WRN-28-10
        & 64.93 $\pm$ 0.18 & \textbf{83.18} $\pm$ 0.11 &       ---        &       --- & 74.81 $\pm$ 0.19 & 87.47 $\pm$ 0.13           \\
        MCT (Inst) \cite{kye2020transductive} & ResNet-12 
        & 65.34 $\pm$ 0.63 & 82.15 $\pm$ 0.45 & 69.66 $\pm$ 0.81 & 85.29 $\pm$ 0.49 & 77.84 $\pm$ 0.64 & 89.11 $\pm$ 0.45  \\
        DeepEMD \cite{zhang2020deepemd} & ResNet-12
        & 65.91 $\pm$ 0.82 & 82.41 $\pm$ 0.56 & 71.16 $\pm$ 0.87 & 86.03 $\pm$ 0.58 &       ---        &       ---          \\
        FEAT \cite{ye2020set} & ResNet-12
        & 66.78 $\pm$ 0.20 & 82.05 $\pm$ 0.14 & 70.80 $\pm$ 0.23 & 84.79 $\pm$ 0.16 &       ---        &       ---        \\
        FEAT+Multi-Proto \cite{multi-pro} & ResNet-12
        & 67.24 $\pm$ 0.58 & 82.51 $\pm$ 0.66 &    ---    &    ---    &       ---        &       ---        \\
        \bottomrule
        ProtoNet* (baseline) & ResNet-12
        & 60.85 $\pm$ 0.69 & 79.50 $\pm$ 0.46 & 67.67 $\pm$ 0.73 & 83.67 $\pm$ 0.55 & 71.65 $\pm$ 0.68 & 86.15 $\pm$ 0.44  \\
        ProtoNet+LLS & ResNet-12 
        & \textbf{68.01} $\pm$ 0.63 & \textbf{83.26} $\pm$ 0.43 & \textbf{72.27} $\pm$ 0.71 & \textbf{86.50} $\pm$ 0.46 & \textbf{78.76} $\pm$ 0.67 & \textbf{89.60} $\pm$ 0.43  \\
        \bottomrule
         \cr 
        \end{tabular}
        \vspace{-10pt}
        \caption{ Average classification performance over 1,000 episodes with 95$\%$ confidence interval. ``---" means that the experiment is unavailable. * means the results is re-implemented.}
        \label{tab:performance_comparison_imagenet}
        }
        \vspace{-10pt}
\end{table*}

\section{Experiments}
\label{sec:experiments}
In this section, we evaluate the proposed local-level strategies (LLS), composed of the comparison with related works (Sec. \ref{sec:SOTA}) and the ablation studies (Sec. \ref{sec:ablation}).

\subsection{Experimental Setup}
\label{sec:experimental_setup}

\noindent\textbf{Datasets}:
We validate our framework on \textit{MiniImageNet} \cite{vinyals2016matching}, \textit{TieredImageNet} \cite{ren2018meta}, and \textit{CIFAR-FS} \cite{bertinetto2018meta} datasets. MiniImageNet and TieredImageNet are subsets of ILSVRC-2012 \cite{russakovsky2015imagenet}, while CIFAR-FS is a variant of CIFAR-100 dataset. They contain 64/16/20, 351/97/160, and 64/16/20 categories for training/validation/test, respectively.

%

\noindent\textbf{Implementation Details}: 
We use Pytorch to implement all our experiments on one NVIDIA TITAN GPU. All input images are resized to $84 \times 84$ and the channel size of each model block output is 64-128-256-512, respectively. In addition, we also adopt horizontal flip and random crop as data augmentation in the training phase. During training, the model is trained by an SGD optimizer for 50,000 episodes, in which $N'$ and $K'$ are set to 15 and 9. For TieredImageNet, we also adopt a batch-based pre-training step with only local-level classification loss followed by the episode-based training, in which the learning rates are initialized to 0.1 and 0.01 with the decay of factor 0.1 at every 20,000 episodes.  For other datasets, the learning rate is initialized to 0.1 and declines to 0.006 and 0.0012 at 30,000 episodes and 45,000 episodes, respectively. The balanced factor in Eq. (\ref{eq:l_all}) is set to $\lambda_S = 0.2, \lambda_R = 0.0001 $. Other hyper-parameters are selected by the dataset validation, resulting in $\gamma=0.6, \beta=0.8$ for MiniImage, $\gamma=0.4, \beta=0.9$ for TieredImageNet, and $\gamma=0.2, \beta=0.5$ for CIFAR-FS.

\subsection{Comparison with the State-of-the-art Methods}
\label{sec:SOTA}
Tab. \ref{tab:performance_comparison_imagenet} shows the comparison with the latest methods. For a fair comparison, we report the model backbone and the average accuracy with 95$\%$ confidence interval over 1,000 episodes. As shown in Tab. \ref{tab:performance_comparison_imagenet}, LLS can significantly boost the performance of the baseline ProtoNet with up 2.8${\%}$-7.2${\%}$ on all benchmark datasets. Compared with competitive FEAT \cite{ye2020set}, Multi-Proto \cite{multi-pro}, and DeepEMD \cite{zhang2020deepemd} adopting extra Transformer module, BERT module, and high-complexity distance metric respectively, we only adopt the simple ProtoNet framework and light-weight strategies, but still achieve the state-of-the-art accuracy on all benchmark datasets. 

\subsection{Ablation Studies}
\label{sec:ablation}
Tab. \ref{tab:ablation_over} shows the ablation studies of three local-level strategies: local-agnostic training (LAT), local-level similarity measure (LSM), and local-level knowledge transfer (LKT). The performance when none of them is adopted can be regarded as our \textbf{baseline}, which is re-implemented from ProtoNet \cite{snell2017prototypical}. As can be seen, the proposed LLS can achieve performance improvements of 3.7${\%}$-7.2${\%}$ on the two datasets. The specific improvements are 2.3${\%}$-4.4${\%}$ by LAT, 0.8${\%}$-2.1${\%}$ by LSM, and 0.4${\%}$-1.3${\%}$ by LKT, respectively.

\begin{table}[h]
    \small
	\setlength{\abovecaptionskip}{0.0cm}
	\setlength{\abovecaptionskip}{0.0cm}
	\setlength{\belowcaptionskip}{0.1cm}
	\setlength{\tabcolsep}{0.8mm}{
		\centering
		\begin{tabular}{cccC{0.9cm}C{0.9cm}C{0.9cm}C{1.35cm}}
			\toprule
			\multirow{2}{*}{LAT} & \multirow{2}{*}{LSM} & \multirow{2}{*}{LKT} & \multicolumn{2}{c}{MiniImageNet} & \multicolumn{2}{c}{CIFAR-FS} \\
			\cline{4-7}
			& & & 1-shot & 5-shot & 1-shot & 5-shot  \\
			\ding{55} & \ding{55} & \ding{55} &  60.85 & 79.50 & 71.65 & 85.45 \\
			Cls       & \ding{55} & \ding{55} &  63.04 & 81.34 & 75.44 & 87.29 \\
			Cls+Reg   & \ding{55} & \ding{55} &  64.69 & 81.82 & 76.05 & 88.21 \\
			Cls+Reg   & Loc       & \ding{55} &  66.19 & 82.30 & 76.40 & 88.99 \\
			Cls+Reg   & Loc+Mat   & \ding{55} &  66.72 & 82.87 & 77.49 & 89.08\\
			Cls+Reg   & Loc+Mat   & \ding{51} &  \textbf{68.01} & \textbf{83.26}     & \textbf{78.76} & \textbf{89.60}     \\
			\bottomrule
			\cr 
		\end{tabular}
    \vspace{-15pt}
	\caption{Ablation studies of three local-level strategies with 1,000 episodes on MiniImageNet and CIFAR-FS.}
    \label{tab:ablation_over}
    \vspace{-15pt}
	}
\end{table}


\section{Conclusion}
In this paper, we focus on the local-level feature and propose a series of local-level strategies to enhance the few-shot image classification by avoiding the discriminative location bias and information loss in local details. Extensive experiments show that the proposed local-level strategies can achieve significant improvements with 2.8$\%$–7.2$\%$ over the baseline, which also achieves the state-of-the-art accuracy. We hope our proposed new perspective on the local-level feature can inspire future works in few-shot learning.


\end{document}